%% file: main.tex
\documentclass{article}

\usepackage[preprint]{Styles/neurips_2024}

\usepackage[colorlinks,citecolor=green]{hyperref}
\usepackage[utf8]{inputenc} 
\usepackage[T1]{fontenc}    
\usepackage{hyperref}       
\usepackage{url}            
\usepackage{booktabs}       
\usepackage{amsfonts}       
\usepackage{nicefrac}       
\usepackage{microtype}      
\usepackage{graphicx}
\input{macros}

\usepackage{algpseudocode}
\usepackage{algorithm}
\usepackage{listings}
\usepackage{multirow}
\usepackage[table]{xcolor}
\usepackage{amsmath}
\usepackage{newtxtext} 
\usepackage{blindtext}
\usepackage{subcaption}
\usepackage{geometry}
\usepackage{xcolor}

\definecolor{limegreen}{rgb}{0.2, 0.8, 0.2}
\definecolor{ered}{rgb}{0.72, 0.16, 0.2}
\definecolor{eblue}{rgb}{0.36, 0.36, 0.36}

\title{Dataset Size Recovery from LoRA Weights}

\author{Mohammad Salama \qquad Jonathan Kahana \qquad Eliahu Horwitz \qquad Yedid Hoshen\\
  School of Computer Science and Engineering\\
  The Hebrew University of Jerusalem, Israel\\
      \small\url{https://vision.huji.ac.il/dsire/}\\
  \small{\texttt{\{ mohammad.salama3, jonathan.kahana, eliahu.horwitz, yedid.hoshen\}@mail.huji.ac.il}}\\
}

\begin{document}
\maketitle
\bibpunct{[}{]}{,}{n}{}{,}
\input{00_abstract}
\input{01_introduction}

\input{02_related_work}
\input{03_preliminaries}
\input{05_method}
\input{06_dataset}
\input{06_experiments}
\input{07_ablation}
\input{10_discussion.tex}
\input{08_coclusion}

\bibliographystyle{plainnat}
\bibliography{references.bib}

\input{09_appendix}
\bibpunct{(}{)}{;}{a}{,}{,}
\end{document}

%% file: macros.tex
\newif\ifdraft
\drafttrue

\ifdraft
\newcommand{\yedid}[1]{{\color{cyan}[\textbf{Yedid:} #1]}}
\newcommand{\salama}[1]{{\color{orange}[\textbf{Salama:} #1]}}

\newcommand{\others}[1]{{\color{green}[\textbf{others:} #1]}}
\newcommand{\eliahu}[1]{{\color{magenta}[\textbf{Eliahu:} #1]}}

\else
\newcommand{\yedid}[1]{}
\newcommand{\salama}[1]{}
\newcommand{\others}[1]{}
\newcommand{\eliahu}[1]{}

\fi

\newcommand{\std}[1]{\textbf{\scriptsize\textcolor{eblue}{$\pm#1$}}} 

%% file: 00_abstract.tex
\begin{abstract}
\label{abstract}

Model inversion and membership inference attacks aim to reconstruct and verify the data which a model was trained on. However, they are not guaranteed to find all training samples as they do not know the size of the training set. In this paper, we introduce a new task: dataset size recovery, that aims to determine the number of samples used to train a model, directly from its weights. We then propose DSiRe, a method for recovering the number of images used to fine-tune a model, in the common case where fine-tuning uses LoRA. We discover that both the norm and the spectrum of the LoRA matrices are closely linked to the fine-tuning dataset size; we leverage this finding to propose a simple yet effective prediction algorithm. To evaluate dataset size recovery of LoRA weights, we develop and release a new benchmark, \textit{LoRA-WiSE}, consisting of over $25\text{,}000$ weight snapshots from more than $2\text{,}000$ diverse LoRA fine-tuned models. Our best classifier can predict the number of fine-tuning images with a mean absolute error of $0.36$ images, establishing the feasibility of this attack.

\end{abstract}

%% file: 01_introduction.tex
\section{Introduction}
\label{sec:intro}
Data is the top factor for the success of machine learning models. Model inversion \citep{fredrikson2015model,yang2019neural,haim2022reconstructing} and membership inference attacks \citep{carlini2022membership,shafran2021membership,jagielski2024students} aim to reconstruct and verify the training data of a model, using its weights\cite{haim2022reconstructing, mia_diffusion_1, mi_weights_1}. While these methods may discover \textit{some} of the training data, they are not guaranteed to recover \textit{all} training samples. One fundamental limit that prevents them from discovering the entirety of the training data is that they do not have a halting condition, as they do not know the size of the training set \citep{haim2022reconstructing}. E.g., in membership inference, the attacker sequentially tests a set of images for membership in the training set but does not know when to halt the algorithm, effectively making its runtime infinite.

Discovering the size of a training dataset given the model weights is important, even without explicit reconstruction of the images themselves. A stock photography provider such as Getty or Shuterstock, may allow users to use their data for fine-tuning personalized generative models and charge them for the number of images actually used for training. Quantifying the dataset size would therefore be essential for billing. Understanding the number of images used to train or fine-tune models is also of great interest to researchers, who wish to understand the costs of replicating a model's performance.

We therefore propose a new task: \textit{Dataset Size Recovery}, which aims to recover the number of training images based on the model's weights. We tackle the important special case of recovering the number of images used to fine-tune a model, where fine-tuning uses Low-Rank Adaption (LoRA)\cite{lora}. 
These LoRA personalized text-to-image foundation models \cite{dreambooth} are among the most commonly trained models, as evident by the success of marketplaces for public sharing of them such as CivitAI and HuggingFace. 

\begin{figure*}[t!]
    \includegraphics[width=\linewidth]{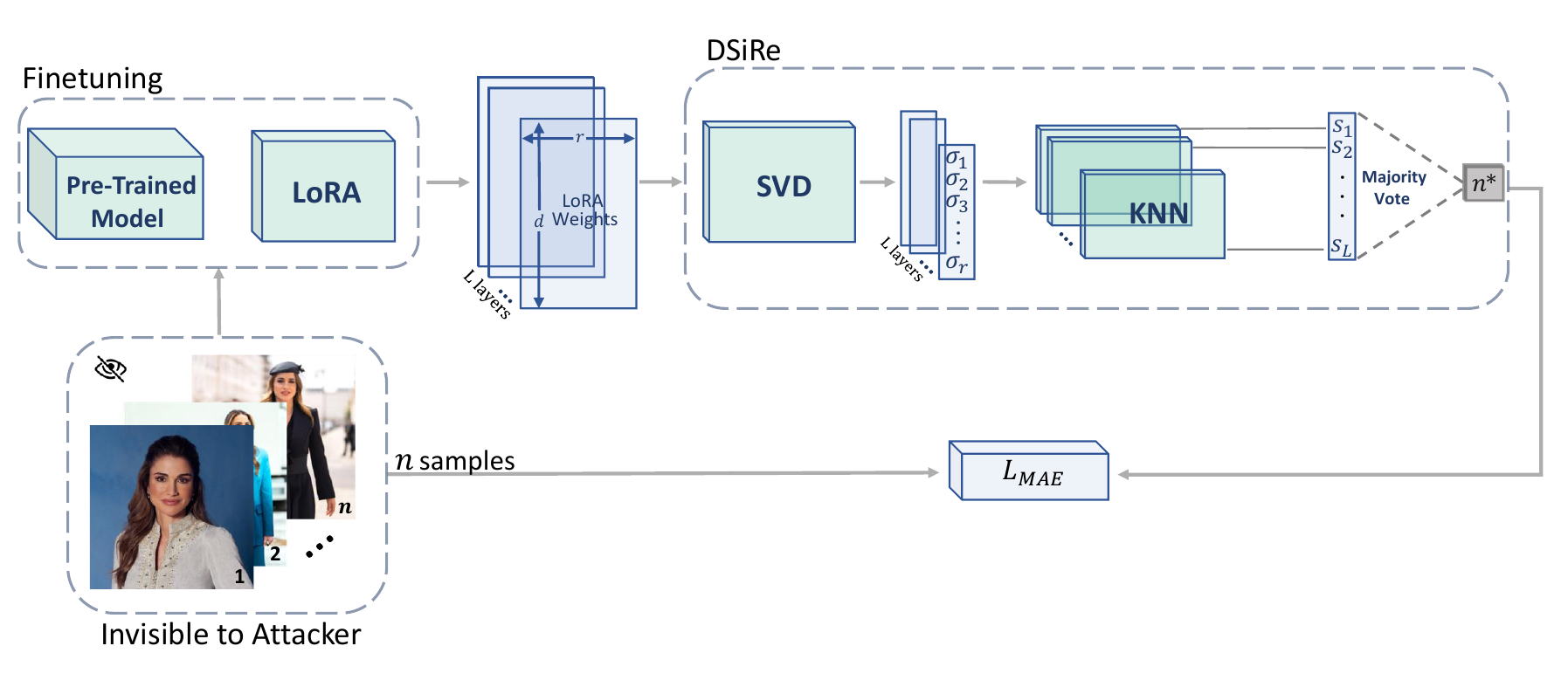}
    \vspace{0.1cm}
    \caption{\textit{\textbf{DSiRe:}} We introduce the task of \textit{dataset size recovery}, which aims to recover the dataset size used to LoRA fine-tune a model based on its weights. DSiRe extracts the singular values of each LoRA matrix and treats them as features. These features are then used to train a set of layer-specific nearest-neighbor classifiers which predict the dataset size.}
    \label{fig:dsire}
\end{figure*}

We first analyze the relationship between LoRA fine-tuning weights and their corresponding dataset size. 
Our observations suggest that the Frobenius norm of LoRA matrices is highly predictive of their fine-tuning dataset size. As a single scalar per weight matrix provides limited expressivity, singular values of each weight matrix serve as more expressive feature vectors. With this in mind, we present \textit{DSiRe} (\textbf{D}ataset \textbf{Si}ze\textbf{ Re}covery). The method recovers the fine-tuning dataset size from the LoRA weight spectrum. DSiRe classifies the dataset size using a trained classifier on top of the weight spectrum of the layer. In practice, we found that a very simple classifier suffices; in our experiments, DSiRe uses the nearest neighbor classifier. For an overview of the method see Fig~\ref{fig:dsire}.

To enable the evaluation of this work and encourage future research, we release a new, large-scale, and diverse dataset: \textit{LoRA-WiSE}. The dataset comprises over $25\text{,}000$ weights checkpoints drawn from more than $2\text{,}000$ independent LoRA models, spanning different dataset sizes, backbones, ranks, and personalization sets. On LoRA-WiSE, DSiRe recovers the dataset sizes from LoRA weights with a Mean Absolute Error (MAE) of $0.36$, demonstrating that our method is highly effective in realistic settings.

To summarize, our main contributions are:
\begin{enumerate}
   \item Introducing the task of dataset size recovery.
   \item Presenting DSiRe, a method for recovering dataset size for LoRA fine-tuning.
   \item Releasing LoRA-WiSE, a comprehensive dataset size recovery evaluation suite based on Stable-Diffusion-fine-tuned LoRAs. 
\end{enumerate}

%% file: 02_related_work.tex
\section{Related work}
\label{sec:related_work}

\subsection{Model Fine-Tuning}
Model fine-tuning \citep{controlnet, lit, spa_text} adapts a model for a downstream task and is considered a cornerstone in machine learning. The emergence of large foundation models \citep{radford2021learning, touvron2023llama, brown2020language, stablediffusion} has made standard fine-tuning costly and unattainable without substantial resources. Parameter-Efficient Fine-Tuning (PEFT) methods were then proposed \citep{lora, qlora, adapters, prefix_tuning, prompt_tuning, p_tuning, unified_peft, ia3, vpt, ada_lora, mpt, fedpara}, offering various ways to fine-tune models with fewer optimized parameters. Among these methods, LoRA \citep{lora} stands out, proposing to train additive low-rank weight matrices while keeping the pre-trained weights frozen. LoRA was found to be very effective across several modalities \citep{prolific_dreamer, mplug, break_a_scene}. Recently, \citet{spectraldetuning} identified a security issue in LoRA fine-tuning, demonstrating that multiple LoRAs can be used to recover the original pre-trained weights. In this paper, we uncover a new use case of LoRA fine-tuning, specifically focusing on the recovery of the dataset size from text-to-image models fine-tuned via LoRA.

\subsection{Membership Inference \& Model Inversion Attacks}
Two privacy vulnerabilities found in machine learning models
are Membership Inference Attack (MIA) \citep{salem2018ml,carlini2022membership,mia_machine,shafran2021membership,jagielski2024students} and Model Inversion \citep{fredrikson2015model,yang2019neural,he2019model,yin2020dreaming,haim2022reconstructing}. First presented by \cite{mia_shokri}, MIAs aim to verify whether a certain image was in the training dataset of a given model. Typically, MIAs assumes that training samples are over-fitted proposing various membership criteria; either by looking for lower loss values \cite{sablayrolles2019white,yeom2018privacy} or some other metrics \cite{watson2021importance,carlini2022membership}. 
In generative models, MIAs have been extensively researched as well \cite{mia_gen, mia_gen_2, mia_gan}, including recent attacks against diffusion models \cite{mia_diffusion_2, mia_diffusion_3}.

Model inversion is a similar attack, in a data-free setting. Introduced by \cite{fredrikson2015model}, model inversion methods wish to generate training samples from scratch, instead of asking whether a known specific image was in the training set. Model inversion is also used for settings where data is unavailable, e.g., data-free quantization \citep{choi2021qimera,xu2020generative,li2023psaq} and data-free distillation \citep{lopes2017data,zhu2021data,zhang2022fine,fang2022up,shao2023data}.

\citet{haim2022reconstructing} emphasized the importance of recovering the training set size for model inversion applications. When this size is unknown, it prevents model inversion attacks from reconstructing the entire dataset a model was trained on, as it is unclear how many samples are sufficient. Our work specifically addresses this issue by uncovering a new vulnerability in fine-tuned models, which enables us to infer the size of the dataset used for fine-tuning.

%% file: 03_preliminaries.tex
\section{Motivation}
\label{sec:pre_analysis}

\subsection{Background: LoRA fine-tuning.} 

Fine-tuning large foundation models can be a computationally expensive process, as it modifies each weight matrix of the pre-trained model $W \in \mathcal{R}^{d \times k}$, by an additive fine-tuning matrix $\Delta W$, as follows:
\begin{equation}
    W^{\prime}=W + \Delta W
\end{equation}

Recently, \citet{lora} introduced Low-Rank Adaptation (LoRA) for efficient fine-tuning. In LoRA, the fine-tuning matrix $\Delta W_i$ is low-rank, i.e., we choose its rank $r$ s.t. $r \ll min(d,k)$. An efficient and SGD amenable way to implement low-rank matrices is to parameterize them as the product of two rectangular matrices so that $B_i \in \mathbb{R}^{d\times r}$, $A_i \in \mathbb{R}^{r\times k}$. The fine-tuning matrix is therefore given by:
\begin{equation}
   \Delta W_i = B_iA_i
\end{equation}

\subsection{Analysing the LoRA Spectrum.}
\label{subsec:analysis}
Our hypothesis is that the difference between pre-fine-tuning and post-fine-tuning weights, denoted as $\Delta W_i$, encodes information about the size of the fine-tuning dataset. Here, we focus on the case where fine-tuning uses LoRA, i.e., where $\Delta W_i$ is low-rank, which is emerging as the most popular fine-tuning paradigm for foundation models. 

\begin{figure}[h!]
    \centering
    \begin{subfigure}[b]{0.495\textwidth}
        \includegraphics[width=\textwidth]{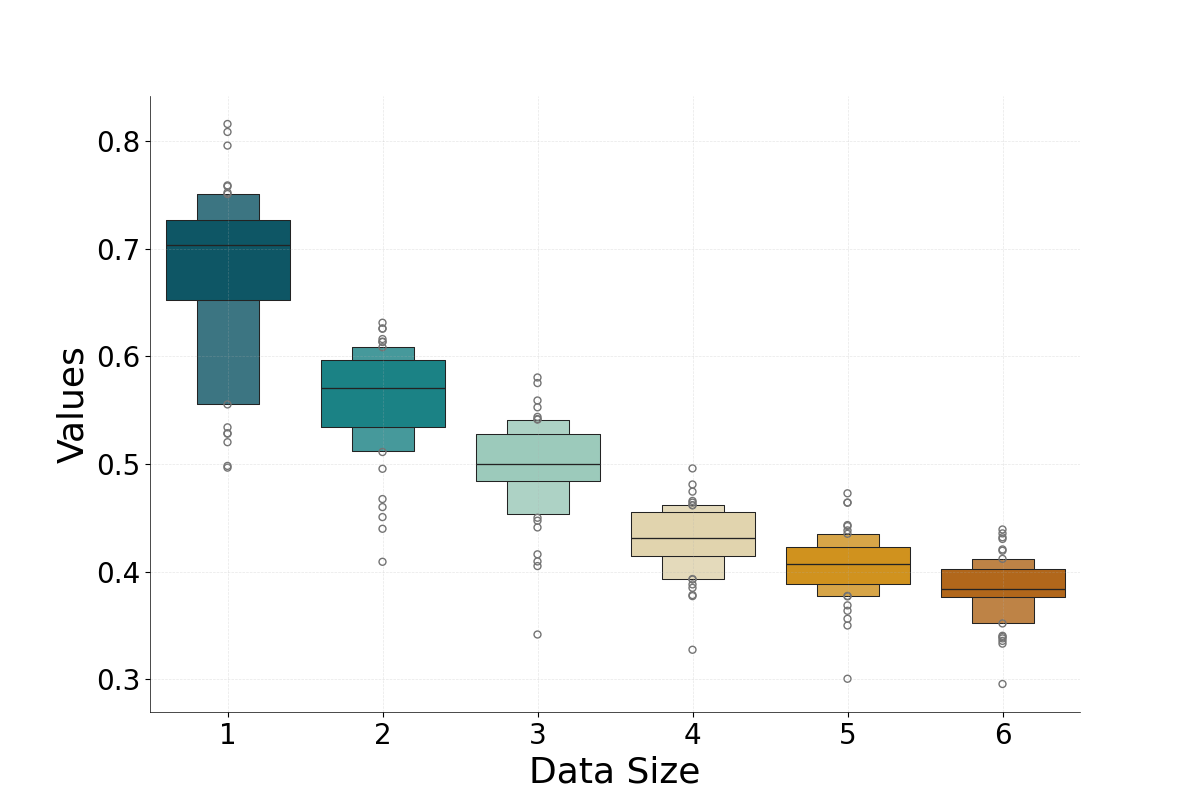}
        \caption{}
        \label{fig:forb_motivation}
    \end{subfigure}
    \begin{subfigure}[b]{0.495\textwidth}
        \includegraphics[width=\textwidth]{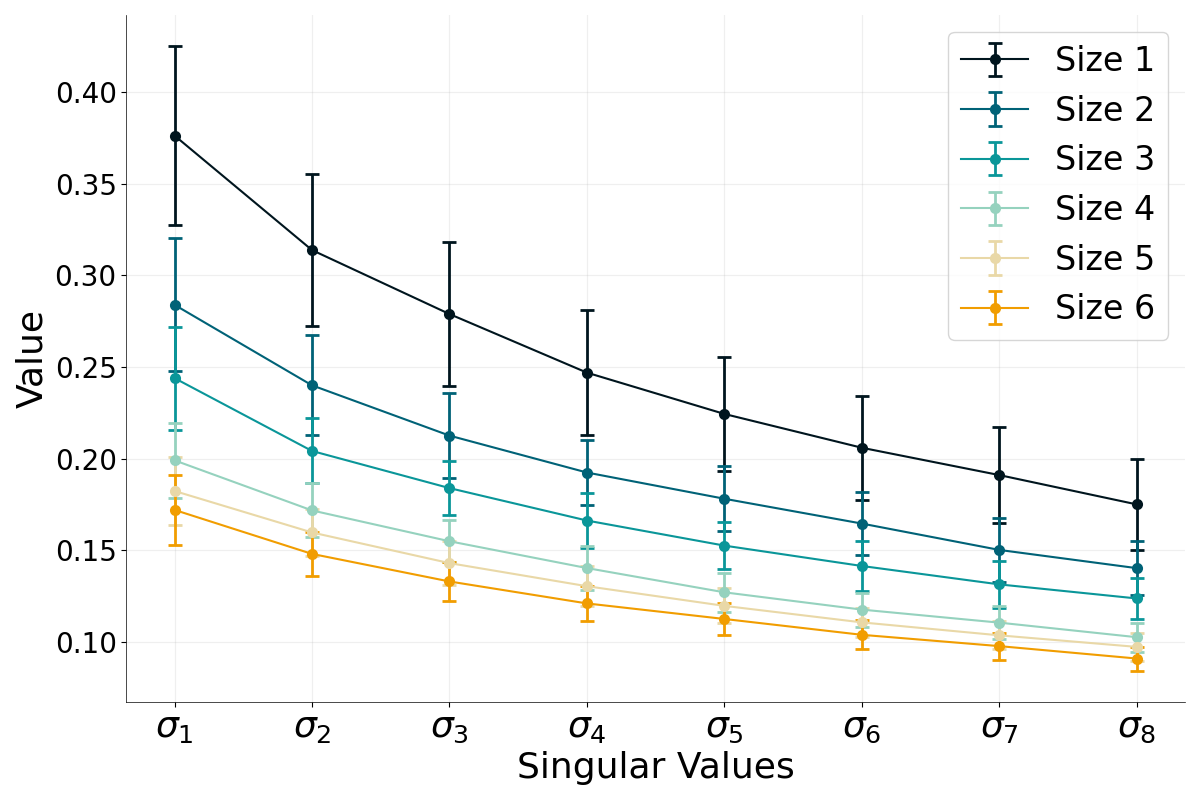}
        \caption{}
        \label{fig:svd_motivation}
    \end{subfigure}
    \caption{\textit{\textbf{Norm and Spectrum of Fine-Tuning Weights vs. Dataset Size.}} Analysis of $210$ Stable Diffusion 1.5 models fine-tuned on datasets of sizes from $1-6$.  (a) Frobenius norm range per dataset size (b) Singular values per dataset size. There is a clear negative correlation between weight/spectrum magnitudes and the size of the fine-tuning dataset.}
    \label{fig:motivatins_analysis}
\end{figure}

We begin by considering a very simple statistic of each fine-tuning matrix, its Frobenius norm that we denote $s_{F}$. 
\begin{equation}
   s_{F} = \sum_{ij} |\Delta W_{ij}|^2
\end{equation}
The norm of a weight matrix correlates with the function complexity the network can express. For example, weight decay, that effectively constrains the norm is a common way for regularizing models. We suggest a simple experiment to analyze the correlation between $s_F$ and the fine-tuning dataset size. We fine-tune Stable Diffusion (SD) 1.5 on $50$ micro-datasets of sizes $1$ to $6$ images, keeping all hyper-parameters, apart from the dataset size, fixed. Fig.~\ref{fig:forb_motivation} shows the range of values of the Frobenius norm statistic $s_F$ for each dataset size. It is clear that $s_F$ is negatively correlated to the dataset size. We motivate this correlation by over-fitting, i.e., models over-fit faster to smaller dataset sizes, leading to larger sizes of $s_F$.

To obtain a deeper understanding, we monitor another statistic of the fine-tuning matrix, its singular values spectrum.  As we consider the case of LoRA fine-tuning, the spectrum of $\Delta W$ has at most $r$ non-zero values (where $r$ is the rank). We denote the $m^{th}$ spectral value as $\sigma_m$, where $m \in [1,\dots,r]$. The range of values of were plotted  $\sigma_m$ across different values of $m$ and different dataset sizes in Fig.~\ref{fig:svd_motivation}. We note there is a better separation between different dataset sizes for the largest singular values. This suggests that the spectrum is more discriminative than the scalar Frobenius norm. Overall, both $s_F$ and the spectrum indicate larger values for small dataset sizes.

\begin{figure}[t]
    \centering
    \begin{subfigure}[b]{0.495\textwidth}
        \includegraphics[width=\textwidth]{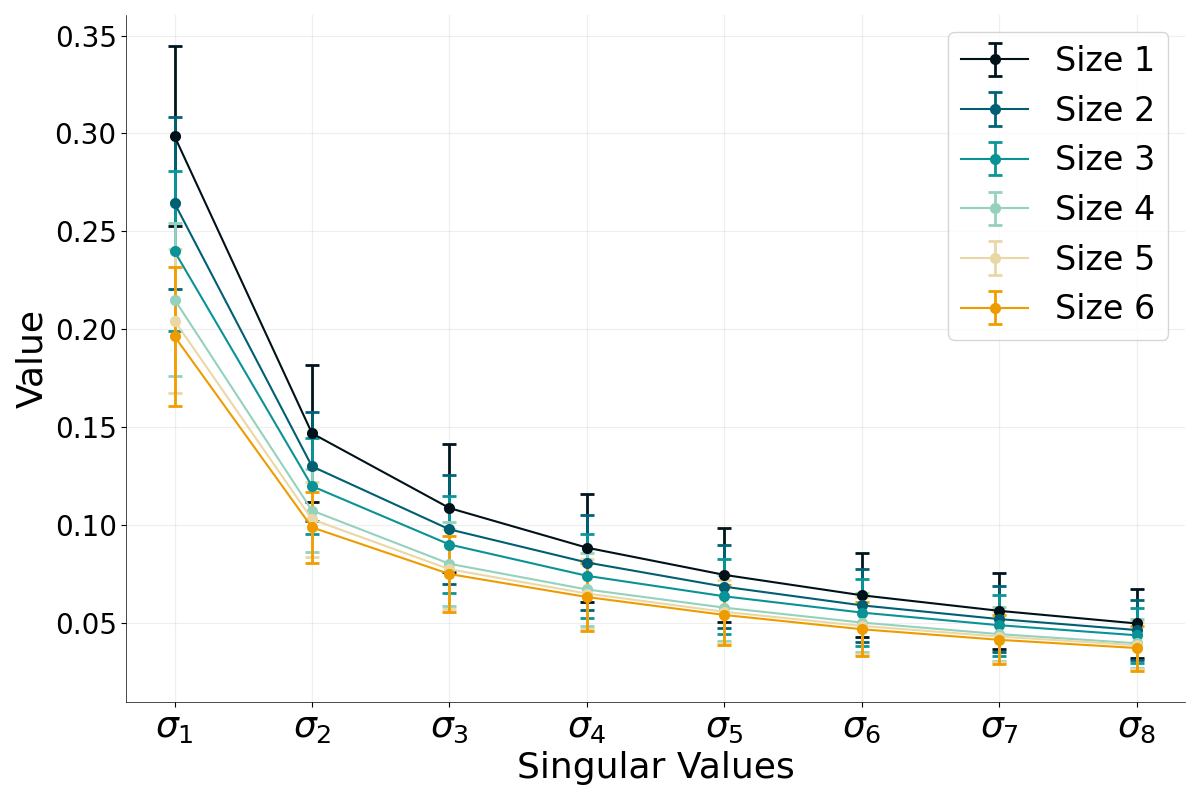}
        \caption{}
        \label{fig:first_down_block}
    \end{subfigure}
    \begin{subfigure}[b]{0.495\textwidth}
        \includegraphics[width=\textwidth]{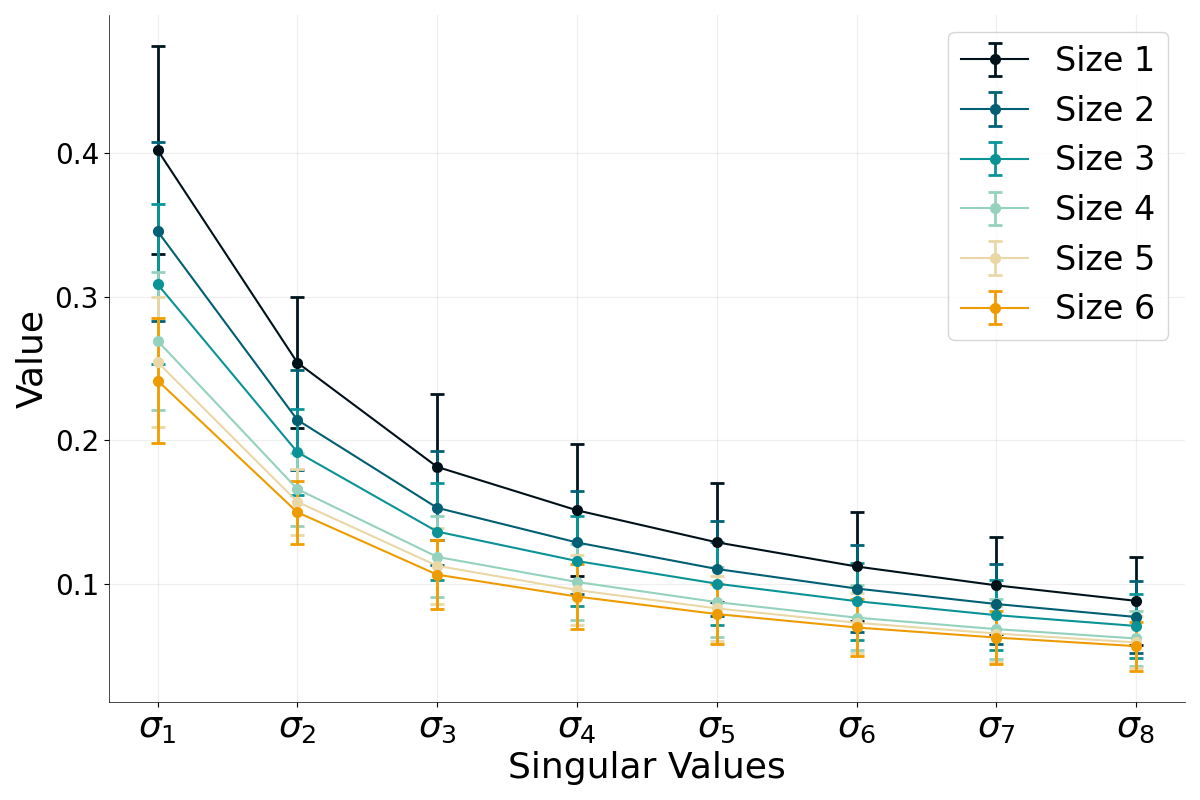}
        \caption{}
        \label{fig:last_upper_block}
    \end{subfigure}
    \caption{\textit{\textbf{Spectrum Ranges of 2 Different Layers.}} Singular values distribution of two layers on opposite sides of Stable Diffusion 1.5 UNet, fine-tuned on datasets of sizes $1-6$. (a) First down block (b) Last upper block. The last upper block shows greater separation of singular values compared to the first down block, highlighting that not all layers are born equally for dataset size recovery.
    }
    \label{fig:motivation_different_blocks}
\end{figure}

Finally, we analyzed how discriminative different layers are for predicting fine-tuning dataset size. We plot the spectra of two different layers - in the first down block and the last up block in Fig.~\ref{fig:motivation_different_blocks}. We can see that the layer is more discriminative than the former one. Indeed,the up layers are found to be more discriminative than the down ones. Without a clear explanation, it can hypothesize that the UNet decoder is more prone to over-fitting than the encoder.  It is noteworthy that our experiments revealed that no single layer is discriminative for all models, suggesting that weighting the results from all layers is better.  

%% file: 05_method.tex
\section{Method}
\subsection{Task Definition: Dataset Size Recovery}
\label{sec:problem_form}

We introduce the task of Dataset Size Recovery for fine-tuned dataset, a new attack vector against fine-tuned models. Formally, given the fine-tuning weights of all layers of a model denoted as $\Delta \mathcal{W} = [\Delta W_1, \Delta W_2 ... \Delta W_L]$, our task is to recover the number of images $n$ that the model was fine-tuned on. More formally, we wish to find a function $f$, such that:
\begin{equation} \label{eq:dsir_task}
    n = f(\Delta \mathcal{W})
\end{equation}
The effectiveness of this attack was measured by the MAE between $f(\Delta \mathcal{W})$ and $n$ across a set of models. 

\subsection{DSiRe}

We propose DSiRe (\textbf{D}ataset \textbf{Si}ze \textbf{Re}covery), a supervised method for recovering dataset size from LoRA fine-tuned weights. Our approach first constructs a training dataset by fine-tuning multiple LoRA models on concept personalization sets across a range of dataset sizes (see Sec. \ref{sec:dataset}). It then trains a predictor function $f$ that acts on a set of fine-tuning weights of each model and outputs the predicted dataset size $n$. At test time, it generalizes to unseen models trained with different concepts. The method can be seen in Fig.~\ref{fig:dsire}

\paragraph{Training set synthesis.} We first synthesize a training set by fine-tuning our model on each of $N_{train}$ datasets, each containing a set of training images. The datasets span a range of sizes; in this paper, we tested the ranges $1-6$,$1-50$, and $1-1000$. The result is a set of $N_{train}$ models $\Delta \mathcal{W}_m$, each with a corresponding label of the dataset size $n_m$.

\paragraph{Predictor training.} Given the set of $N_{train}$ labeled models, we wish to train a predictor that maps the fine-tuning weights $\mathcal{W}_m$ to dataset size $n_m$. Motivated by the results of our analysis (see Sec.~\ref{sec:pre_analysis}), we describe the weights of each fine-tuning layer using its spectrum $\Sigma$ consisting of $r$ singular values. Let us define the features of each model as the set of spectra of all its $L$ layers $f = [\Sigma_1,\Sigma_2..\Sigma_L]$. We tested many different predictors and ablated them in Tab.~\ref{tab:predictors_results}. Overall, the simple Nearest Neighbor (NN) ensemble performed the best. During inference, given a new model, for each layer, we retrieve the NN layer that has the most similar spectrum to the layer of the target model. Each layer votes for the dataset size of its NN layer. The overall prediction is the dataset size that most layers voted for. 

%% file: 06_dataset.tex
\section{LoRA WiSE Benchmark}
\label{sec:dataset}
We present the LoRA Weight Size Evaluation (LoRA-WiSE) benchmark, a comprehensive benchmark specifically designed to evaluate LoRA dataset size recovery methods, for generative models. More specifically, it features the weights of $2350$ Stable Diffusion \cite{stablediffusion} models, which were LoRA fine-tuned by a standard, popular protocol \citep{dreambooth,dreambooth_readme}.  Our benchmark includes versions $1.5$ and $2$ of Stable Diffusion, having $2050$ and $300$ trained models for each version respectively.

We fine-tune the models using three different ranges of dataset size: (i) Low data range: $1-6$ images. (ii) Medium data range: $1-50$ images. (iii) High data range: $1-1000$. For each range, we use a discrete set of fine-tuning dataset sizes. In the low and medium ranges, we also provide other versions of these benchmarks with different LoRA ranks and backbones. See Tab.\ref{tab:bench_details} for the precise benchmark details. 

\begin{table}[t!]
    \caption{\textit{\textbf{LoRA WiSE Benchmark Overview.}} The dataset comprises over $25\text{,}000$ weights checkpoints drawn from more than 2000 independent LoRA models, spanning different dataset sizes, backbones, ranks, and personalization sets.}
    \vspace{0.2cm} 
    \centering
    \begin{tabular}{c@{\hskip5pt}c@{\hskip5pt}c@{\hskip5pt}c@{\hskip5pt}c@{\hskip5pt}c@{\hskip5pt}}
    
    Data Range & Dataset Sizes & Source & Backbone & LoRA Rank & $\#$ of Models \\
    \midrule
    \multirow{3}{*}{Low} & \multirow{3}{*}{$1,2,3,4,5,6$} & \multirow{3}{*}{Concept101} & \multirow{3}{*}{SD 1.5} & $8$ & $300$ \\
    &  &  & & $16$ & $300$ \\
    &  &  & & $32$ & $300$ \\
    \midrule
    \multirow{4}{*}{Medium} & \multirow{4}{*}{$1,10,20,30,40,50$} & \multirow{4}{*}{ImageNet} & \multirow{3}{*}{SD 1.5} & $16$ & $300$ \\
    &  &  & & $32$ & $300$ \\
    &  &  & & $64$ & $300$ \\ 
    \cmidrule{4-6}
    &  &  & SD 2 & $32$ & $300$ \\
    \midrule
    High & $1,50,100,500,1000$ & ImageNet & SD 1.5 & $32$ & $250$ \\
    \end{tabular}
        \label{tab:bench_details}
\end{table}

For our low data range set, we choose Concept101 \citep{dataset_multi}, a previously collected set of micro-datasets ($3-15$ images) designed for personalization research. For our medium and high data ranges we use different classes of ImageNet \cite{imagenet} as the data source.
This selection of datasets aims to ensure that the fine-tuned models are drawn from a diverse set of concepts, spanning various categories. 

Each micro-dataset is used to fine-tune the models for each dataset size. The images are randomly selected from the micro-dataset. Each Stable Diffusion model consists of $132$ adapted layers (pairs of $A_i,B_i$), including various layer types, such as self-attention, cross-attention, and MLPs. We save $A_i$, $B_i$ separately, i.e., each model provides a total of $264$ unique weight matrices. We then split each range of this new benchmark (low, medium, and high ranges) into train and test sets based on the micro-datasets. for more details see appendix~\ref{appendix: implementations details} 

%% file: 06_experiments.tex
\section{Experiments}
\label{sec:experiments}
\subsection{Experimental Setup} We evaluate DSiRe on the LoRA-WiSE benchmark. Unless mentioned otherwise, we use  Stable Diffusion 1.5 as the pre-trained model, which we fine-tune using a LoRA of rank 32 (denoted as above $\Delta \mathcal{W}$).
As for the training set, we use $15$ different personalization sets, to train each $\Delta \mathcal{W}$ model, resulting in a training set of $90$ weight samples for low and medium dataset sizes, and $75$ weight samples for our higher ($1-1000$) range. We evaluate DSiRe using $35$ additional personalization sets. We note, 
We repeat this experiment $10$ times, including subset sampling. The reported performance metrics are
the average and standard deviation over the experiments.

\paragraph{Baseline.} We compare DSiRe to a baseline, denoted as Frobenius-NN, which predicts the dataset size using a nearest neighbor classifier on top of the Frobenius norms of the layers' LoRA weights. Similar to DSiRe, the Frobenius-NN is fitted separately to each layer, and then a majority vote rule is applied to select the prediction from all layer-wise predictions. The analysis in Sec \ref{sec:pre_analysis} provides motivation for this baseline.

\paragraph{Evaluation metrics.} As described in Sec. \ref{sec:problem_form}, our main evaluation metric is Mean Absolute Error (MAE). For completeness, we choose to report two complementary metrics as well: (i) Accuracy. (ii) Mean Absolute Percentage Error (MAPE). Since DSiRe predicts dataset sizes, simple accuracy does not adequately measure its effectiveness, e.g., predicting $4$ when the true value is $5$ is not as bad as predicting $1$. We therefore provide MAPE scores as well, which compute the percentile from the ground truth that is equal to the absolute error.

\subsection{Results}
We begin by evaluating on $1-6$ fine-tuning images. When using both the singular values as features for DSiRe and the Frobenius norm for Frobenius-NN, we find that they yield relatively successful results in recovering the dataset size.

These low data range results are presented in Table \ref{tab:range_performances}. As DSiRe outperforms Frobenius-NN by a small margin ($>3\%$),  we conclude that the results align with our analysis (see Sec.~\ref{subsec:analysis}), which demonstrate that both singular values and Frobenius norm are indeed predictive of the dataset size.

Mid-range fine-tuning dataset sizes ($10-50$ images) are common in artistic LoRA fine-tuning. We therefore present the results of our method using $1-50$ fine-tuning images in Table \ref{tab:range_performances}, showing that DSiRe performs well with an MAE of $1.48$. In this data range, the Frobenius-NN baseline achieves comparable results to DSiRe across all metrics, demonstrating good performance. While the absolute MAE value is larger than in the low data range case, it is relatively small compared to the range of data sizes. The accuracy and mean absolute percentage error (MAPE) scores of both methods further support this observation. Fig. \ref{fig:conf_mat} shows another favorable property of our approach: its mistakes are usually near hits, i.e., large errors between ground truth and predicted labels are rare.

\begin{table}[t!]
    \centering
    \caption{\textit{\textbf{Performance Comparison of Dataset Size Recovery Methods across Different Ranges.}} Performance of Frobenius-NN and DSiRe across different data ranges $(1-6,\space1-50,\space1-1000)$ using Stable Diffusion 1.5. These decent results aligns with our analysis (see Sec~\ref{sec:pre_analysis}), showed that both SVD and Frobenius norm are effective features for dataset size recovery. However, DSiRe outperfomrs the Frobenius-NN on all evaluation metrics.}
    \label{tab:range_performances}
    \vspace{0.2cm}
    \begin{tabular}{c@{\hskip5pt}lccc}    
        Data Range & Method & MAE $\downarrow$ & MAPE$(\%) \downarrow$ & Acc$(\%) \uparrow$\\
        \toprule
        \multirow{2}{*}{1-6}   & Frobenius-NN & 0.43 \std{0.04} & 15.14 \std{2.12} & 65.29 \std{2.42} \\
                               & DSiRe          & \textbf{0.36} \textbf{\std{0.04}} & \textbf{11.36} \textbf{\std{1.55}} & \textbf{69.30} \textbf{\std{3.83}} \\
        \midrule
        \multirow{2}{*}{1-50}  & Frobenius-NN & 1.56 \std{0.19} & 4.16 \std{0.75} & 85.33 \std{1.81} \\
                               & DSiRe          & \textbf{1.48} \textbf{\std{0.21}} & \textbf{3.97} \textbf{\std{0.73}} & \textbf{86.10} \textbf{\std{1.99}} \\
        \midrule
        \multirow{2}{*}{1-1000}& Frobenius-NN & 68.62 \std{5.53} & 9.25 \std{1.21} & 86.51 \std{1.12} \\
                               & DSiRe          &\textbf{ 41.77} \std{6.61} & \textbf{5.96} \std{1.46} & \textbf{91.90} \std{1.28} \\
    \end{tabular}
\end{table}

At larger data quantities, dataset size recovery could aid in better understanding data collection quantities needed for fine-tuning. Therefore, we conducted an additional experiment using models trained with higher data ranges, having $1,50,100,500$ and $1000$ image samples per model (note that here we have $5$ dataset size classes). Results, presented in Tab. \ref{tab:range_performances}, shows DSiRe is able to detect the dataset size with more than $90\%$ accuracy, and a MAPE score of only $6\%$. Additionally, in Fig. \ref{fig:high_conf_matrix} we show the confusion matrix generated by DSiRe, where we see that most of the errors happen between adjacent classes.

\paragraph{Other Backbone} The LoRA fine-tuning technique is commonly used by popular text-to-image models. A desirable aspect of our paradigm is being robust to model architecture. In this part, We test the robustness of DSiRe to the backbone model by evaluating it on Stable Diffusion 2.0. We note that these models do not share pre-training weights, as Stable Diffusion 2.0 was \emph{not} fine-tuned from a previous version.
Tab. \ref{table:different_sd} shows that DSiRE performs well on Stable Diffusion 2.0, reaching around $77\%$ accuracy. This provides evidence for the correlation between the singular values and dataset size is not specific to one backbone alone.  

\begin{table}[t]
    \caption{\textit{\textbf{Robustness of Dataset Size Recovery Methods on Stable Diffusion 2.0.}} DSiRe recovers dataset size more effectively than Frobenius-NN for the medium data range $(1-50)$ using Stable Diffusion 2.0. This supports the benefit from a more expressive representation given by the singular values, independent to the specific backbone model used.} 
    \vspace{0.2cm}
    \centering
    \begin{tabular}{c@{\hskip5pt}c@{\hskip5pt}c@{\hskip5pt} c@{\hskip5pt}}    
    Method & MAE$\downarrow$ & MAPE$(\%)\downarrow$ & Acc$(\%)\uparrow$\\
    \toprule
    Frobenius-NN  & 2.95 \std{0.28} & 11.99 \std{3.93} & 73.90 \std{2.21} \\
    DSiRe &  2.51 \std{0.22} & 7.46 \std{0.95} & 77.43 \std{1.70} \\
    \end{tabular}
    \label{table:different_sd}
\end{table}

%% file: 07_ablation.tex
\section{Ablation studies}
\label{sec:ablation}

\subsection{Number of Micro-Datasets}
\label{ablation:num_models}
\begin{figure*}[t!]
    \centering
    \begin{minipage}{0.49\textwidth}
        \centering
        \includegraphics[width=\linewidth]{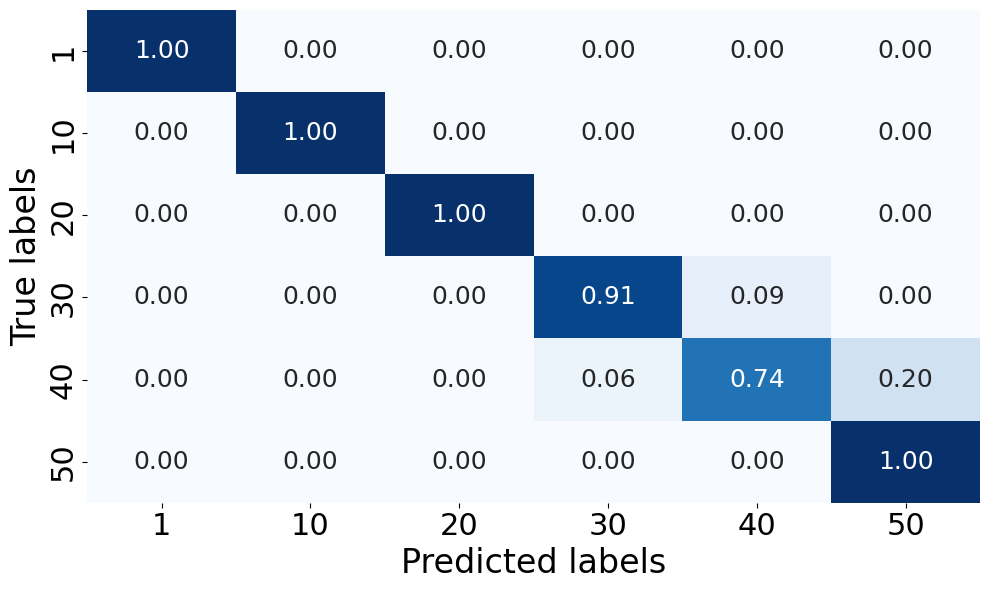}
        \caption{\textit{\textbf{DSiRe Confusion Matrix for Medium Data Range in a single experiment.}} Illustrating DSiRes accuracy in the range of $1-50$ samples, shows that most of the errors are near misses, highlighting DSiRe's precision in dataset size recovery.}
        \label{fig:conf_mat}
    \end{minipage}\hfill
    \begin{minipage}{0.49\textwidth}
        \centering
        \includegraphics[width=\linewidth]{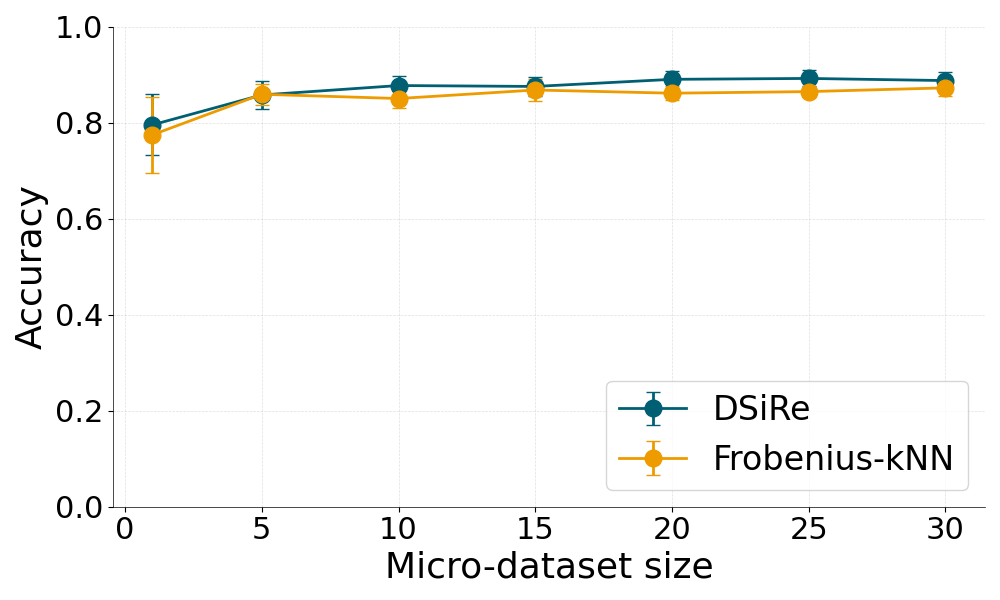}
        \caption{\textit{\textbf{DSiRes Micro-Dataset Size vs. Accuracy, reported on medium data size range $(1-50)$.}} Even a single micro-dataset is sufficient for DSiRe to reach $80\%$ accuracy. This demonstrates its effectiveness with limited training data.}
        
        \label{fig:tr_size}
    \end{minipage}\
\end{figure*}

While our attack is data driven and requires access to the pre-trained model, we find that only a few examples are needed for DSiRe to perform well. E.g., in our medium data size range, our model can reach $86.4\%$ accuracy using only $5$ micro-datasets for training. The full results, presented in Fig.~\ref{fig:tr_size}, showcases that while more samples (fine-tuned models) improves the accuracy of our predictor, even a single micro-dataset is sufficient to achieve around $80\%$ accuracy. This shows that our method is robust to the number of micro-datasets used, even to very small numbers.

\subsection{Robustness to LoRA Hyper-Parameters}
While DSiRe performs well in our settings, it is important to show that it remains robust to hyperparameters. In this section, we will ablate our method using the LoRA-WiSE benchmark by testing DSiRe with models fine-tuned using other recipes. We check two common variations of the LoRA models: (1) LoRA rank (2) seeding. For additional ablation studies on the training steps, batch size, and used classifier see appendix. \ref{sec:appe_lora_parameters}.

\textit{LoRA Rank.} Starting with the LoRa rank ablation, we train DSiRe for different varying LoRA ranks. Tab. \ref{tab:different_ranks} shows the results for medium and low data ranges. Our method is robust to the LoRA rank, achieving similar results in all $3$ tested ranks for both ranges.

\begin{table}[t!]
        \centering
        \captionof{table}{\textit{\textbf{DSiRe Performance with Different LoRA Ranks.}} Desire consistently achieves high accuracy across both low and medium ranges, indicating its robustness regardless of LoRA rank variations.}
        \vspace{0.2cm}
        \begin{tabular}{c c c c c}
            Data Range & LoRA Rank & MAE $\downarrow$ & MAPE$(\%) \downarrow$ & Acc$(\%) \uparrow$\\
            \toprule
            \multirow{3}{*}{$1-6$} & 8  & 0.43 \std{0.04} & 14.8 \std{2.3} & 66 \std{3.08} \\
            & 16 & 0.42 \std{0.03} & 12.4 \std{1.11} & 67.7 \std{2.3} \\
            & 32 & 0.36 \std{0.04} & 11.36 \std{1.55} & 69.30 \std{3.83} \\ 
            \cmidrule{1-5}
            \multirow{3}{*}{$1-50$} & 16  & 1.67 \std{0.17} & 4.32 \std{0.46} & 84.04 \std{1.85} \\
            & 32 & 1.48 \std{0.21} & 3.97 \std{0.73} & 86.10 \std{1.99} \\
            & 64 & 1.41 \std{0.39} & 3.90 \std{1.30} & 86.58 \std{3.45} \\ 
        \end{tabular}
        \label{tab:different_ranks}
\end{table}

\textit{Seeding.} While in the standard recipe, all models use $seed=0$, we also tested the case where all seeds were selected randomly. Tab. \ref{tab:parameters_ablation} shows that the variation in seeds only reduces accuracy by around $4\%$, and that MAE decreases by less than $0.5$. This is not a small change, given that the gap between possible dataset size values is $10$.

\subsection{Choice of Classifier}
\label{subsec:ablate_predictor}
We tested various predictors, including parametric and non-parametric classifiers, as shown in Table~\ref{tab:predictors_results}. Except the NN-full model baseline, our pipeline remains unchanged, all classifiers are fitted separately to each layer and a majority vote rule selects the label from all layer-wise predictions. In contrast, the NN-full model uses a kNN classifier which is fitted to all the layers simultaneously. The results show that the choice of classifier affects the performance significantly. Furthermore, these results confirm our hypothesis from Sec. \ref{sec:pre_analysis}: while each layer is predictive of the dataset size, the accuracy of individual layers is not sufficient. As the layer-wise NN predictor can pool information across layers, it helps DSiRe perform better than the other methods.

%% file: 10_discussion.tex
\section{Discussion}
\label{sec:discussion}

\paragraph{Performance at Low Data Ranges.} While our approach shows promising results, there is room for improvement in lower data regimes, where DSiRe reaches less than $80\%$ accuracy. Improving these results will provide tighter upper bounds for membership inference and model inversion attacks.

\paragraph{Data Driven Solution.} Our method is data driven as it requires training multiple models from each dataset size. However, our analysis shows there is correlation between the Frobenius norm and dataset size (see Fig. \ref{fig:forb_motivation}). This insight could be a stepping stone in developing a data-free solution.

\paragraph{Pre-training dataset size recovery.}
Another interesting application of dataset size recovery is for pre-training cases. Lower bounding the required number of training set samples for foundation models will have a substantial impact on the research community. Answering this question would require scaling up our method to much larger dataset sizes and weight matrix dimensions.

\paragraph{Defense against DSiRe.}
Data augmentation aims to artificially increase the size of the data set by generating more image transformations, such as flipping and rotation on existing training images. Motivated by this observation, we employ data augmentation during the fine-tuning process to inflate the actual number of unique images used for fine-tuning, adding more variation to every image in the dataset. To avoid hurting model convergence during fine-tuning or the quality of the generated images, we applied only the flip augmentation technique to each image in the dataset. This process reduced the correlation between the singular values and the true number of images, as illustrated in Fig. ~\ref{tab:defence_results}.
\begin{table}[t]
    \caption{\textit{\textbf{Performance of various predictors, medium dataset size range $(1-50)$.}} DSiRe with majority voting performs best by combining predictions  from multiple layers, opposed to the nearest neighbor - full model baseline, which uses the spectra of all layers together as features for a single prediction.}
    \centering
    \vspace{0.2cm}
    \begin{tabular}{l@{\hskip5pt}c@{\hskip5pt}c@{\hskip5pt}c@{\hskip5pt}}    
             Model  & MAE $\downarrow$ & MAPE$(\%) \downarrow$ & Acc$(\%) \uparrow$\\
            \midrule
            NN - full model & 7.33 \std{0.75} & 33.37 \std{7.71} & 48.61 \std{3.53} \\
            Ridge Regression & 8.05 \std{0.08} & 38.95 \std{6.24} & 37.09 \std{0.63} \\
            GDA & 2.89 \std{0.48} & 7.87 \std{1.30} & 74.14 \std{3.65} \\
            DSiRe - average vote & 3.57 \std{0.36} & 19.05 \std{3.37} & 64.62 \std{3.58} \\
            DSiRe - majority vote & \textbf{1.48 \std{0.21}} & \textbf{3.97 \std{0.73}} & \textbf{86.10 \std{1.99}}  \\
        \end{tabular}
        \label{tab:predictors_results}
\end{table}
\begin{table}[t!]
    \caption{\textit{\textbf{Defense against DSiRe, on medium data range $(1-50)$.}} We show how the simplest data augmentation (horizontal flipping), can harm DSiRes performance, reducing its accuracy by more than $26\%$.}
    \centering
    \vspace{0.2cm}
    \begin{tabular}{l@{\hskip5pt}c@{\hskip5pt}c@{\hskip5pt}c@{\hskip5pt}}    
    Ablation  & MAE $\downarrow$ & MAPE$(\%) \downarrow$ & Acc$(\%) \uparrow$\\
    \toprule
    Defense & 4.62 \std{0.83} & 17.62 \std{4.82} & 59.71 \std{5.88} \\
    DSiRe & 1.48 \std{0.21} & 3.97 \std{0.73} & 86.10 \std{1.99} \\
    \end{tabular}
    \label{tab:defence_results}
\end{table}

\section{Social impact}
\label{sec:social_impact}
We believe our newly proposed task can positively impact both the research and digital arts communities. When detecting small and mid range dataset sizes, our motivation is to establish an upper bound for membership inference attacks, promoting privacy aware deployment of LoRA models. For larger dataset sizes, it will increase the information on the resources requires to successfully train models. This is often for researchers that need to collect expensive datasets for new fine-tuning tasks e.g., \cite{objectdrop} and \cite{emu}.

%% file: 08_coclusion.tex
\section{Conclusion}
\label{sec:colclusion}
We introduced the new task of dataset size recovery and proposed a method, DSiRe, for learning a predictor for this task for models that use LoRA fine-tuning. Our method showed promising results on a new large-scale dataset that we released. We believe our work can provide an upper bound for model inversion and membership inference attacks. It can also provide both researchers and stock photography owners with a quantitative analytic estimating the data cost for fine-tuning models.
\newpage

%% file: 09_appendix.tex
\newpage
\section{Appendix}
\subsection{Additional Ablation Studies}
We provide more ablation studies of our method. Specifically, we test the training steps, batch size, used classifier type and used LoRA matrices.

\subsection{Robustness to LoRA Hyper-Parameters}
\label{sec:appe_lora_parameters}

\paragraph{Batch Size.} We ablate the batch size, results at shown in Tab. \ref{tab:parameters_ablation}. Despite the change in batch size, DSiRe demonstrates robust performance, achieving a MAE score of $1.94$ compared to the original $1.48$. Additionally, the accuracy only decreases by less than $5\%$, indicating that our method maintains comparable effectiveness even with different batch sizes.

\begin{table}[h!]
    \caption{DSiRe performance using different LoRA hyper-parameters. Medium data range}
    \centering
    \begin{tabular}{l@{\hskip5pt}c@{\hskip5pt}c@{\hskip5pt}c@{\hskip5pt} c@{\hskip5pt}}    
    Ablation  & MAE$\downarrow$ & MAPE$(\%)\downarrow$ & Acc$(\%)\uparrow$\\
    \toprule
    Batch size & 1.94 \std{0.26} & 9.35 \std{1.34} & 81.50 \std{2.55} \\
    Baseline & 1.48 \std{0.21} & 3.97 \std{0.73} & 86.10 \std{1.99} \\
    \end{tabular}
    \label{tab:parameters_ablation}
\end{table}

\paragraph{Training Steps.} To train DSiRe, we first fine-tune a set of LoRA models. These models follow a certain recipe, with a specific amount of training steps. To evaluate robustness, we tested DSiRe on models fine-tuned at different steps, with $1200$ steps as our baseline. As shown in Tab \ref{tab:different_steps}, DSiRe consistintly achieves comparable results across different fintuning steps. e.g. the MAE score ranges from  $2.43$ at 300 steps to $1.40$ at 1400 steps, with accuracy variations within $10\%$.
\begin{table}[h!]
    \caption{\textit{\textbf{DSiRe performance on different checkpoints of Stable Diffusion 1.5 rank 16 range $1-50$}}}
    \centering
    \begin{tabular}{l@{\hskip5pt}c@{\hskip5pt}c@{\hskip5pt}c@{\hskip5pt}}
    \#Steps& MAE$\downarrow$ & MAPE$(\%)\downarrow$ & Acc$(\%)\uparrow$\\
    \toprule
    300  & 2.43 \std{0.20} & 6.82 \std{0.78} & 77.90 \std{1.49} \\
    400  & 2.39 \std{0.20} & 6.72 \std{0.76} & 78.38 \std{1.49} \\
    500  & 2.05 \std{0.15} & 5.55 \std{0.59} & 81.33 \std{1.60} \\
    600  & 1.86 \std{0.10} & 4.59 \std{0.34} & 82.76 \std{0.86} \\
    700  & 1.89 \std{0.21} & 5.13 \std{0.77} & 82.00 \std{2.01} \\
    800  & 1.71 \std{0.29} & 4.59 \std{0.89} & 83.67 \std{2.68} \\
    900  & 1.60 \std{0.22} & 4.21 \std{0.69} & 85.14 \std{2.04} \\
    1000 & 1.62 \std{0.21} & 4.69 \std{0.70} & 85.10 \std{1.77} \\
    1100 & 1.58 \std{0.19} & 4.50 \std{0.90} & 84.48 \std{1.32} \\
    1200 & 1.48 \std{0.21} & 3.97 \std{0.73} & 86.10 \std{1.99} \\
    1300 & 1.46 \std{0.15} & 3.84 \std{0.51} & 86.29 \std{1.55} \\
    1400 & 1.40 \std{0.20} & 3.73 \std{0.76} & 86.76 \std{2.08} \\
    \end{tabular}
    \label{tab:different_steps}
\end{table}

\subsection{Choice of LoRA Matrices}
Seeing in Sec. \ref{sec:pre_analysis} that not all layers are similar in behavior, we test to see if different LoRA matrices also capture different information. In Tab. \ref{tab:lora_layers}, we find that indeed different LoRA matrices capture different information, and lead to substantially other performances. Unsurprisingly, we also find that using all the LoRA matrices combined yields the best result.

\begin{table}[h!]
    \caption{\textit{\textbf{DSiRe performance on different layers of LoRA of the UNet in Stable Diffusion 1.5, range $1-50$:}} }
    \centering
    \begin{tabular}{l@{\hskip5pt}c@{\hskip5pt}c@{\hskip5pt}c@{\hskip5pt} c@{\hskip5pt}}    
            \# Layer Type  & $MAE$ $\downarrow$\ & $MAPE$(\%) $\downarrow$\ & $Acc$(\%)\\
            \midrule
            A & 1.9 \std{0.29} & 5.63 \std{1.26} & 82.52 \std{2.20} \\
            B & 1.57 \std{0.19} & 4.07 \std{0.65} & 84.90 \std{2.09} \\
            BA & 1.61 \std{0.16} & 4.22 \std{0.47} & 85.00 \std{1.45}  \\  
            full model & 1.48 \std{0.21} & 3.97 \std{0.73} & 86.10 \std{1.99}  \\
        \end{tabular}
        \label{tab:lora_layers}
\end{table}

\section{Higher Data Regimes Analysis}

To better understand the results on higher data regimes we provide here the confusion matrix of DSiReusing $1-1000$ training samples. We can see that most of the errors are in larger data classes.

\begin{center}
\begin{figure*}[h!]
\includegraphics[width=0.9\linewidth]{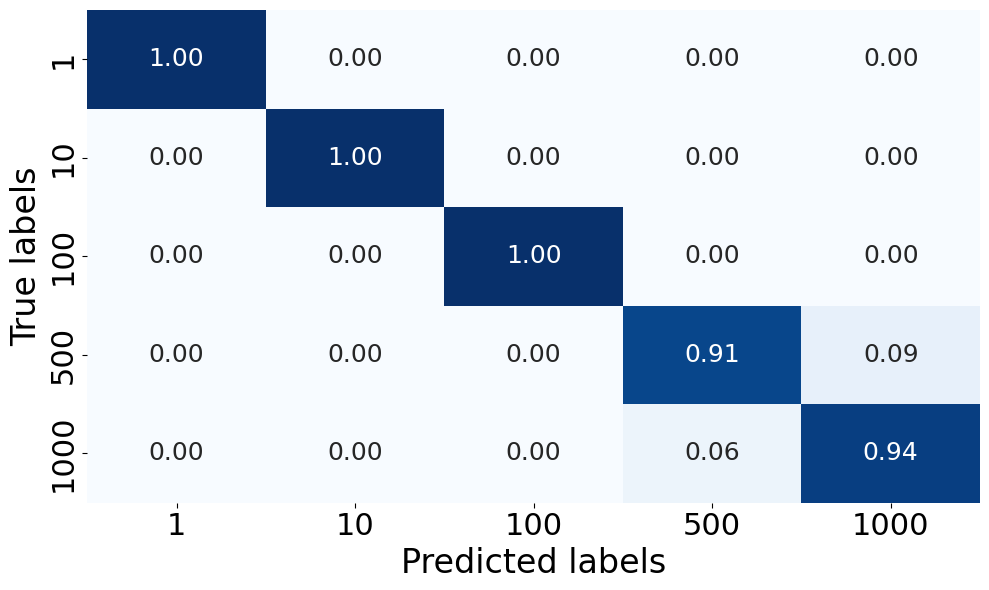}
\caption{\textbf{DSiRe Confusion matrix in High data regime.} Illustrating DSiRe's accuracy in the range data size $(1-1000)$ for a single experiment, showing that most predictions are correct or near misses, highlighting the DSiRe's precision in dataset size recovery.}
\label{fig:high_conf_matrix}
\end{figure*}
\end{center}

\subsection{Implementations details}
\label{appendix: implementations details}
\subsubsection{LoRA-WiSE.}
we now elaborate the implementations details of the LoRA-WiSE bench dataset.
\paragraph{Datasets in all ranges $1-6$, $1-50$, $1-1000$.}

As the Pre-Ft models we use \href{https://huggingface.co/runwayml/stable-diffusion-v1-5/}{\texttt{runwayml/stable-diffusion-v1-5}} and \href{https://huggingface.co/stabilityai/stable-diffusion-2/}{\texttt{stabilityai/stable-diffusion-2}} \cite{stablediffusion}. We fine-tune the models using the PEFT library \cite{peft}. We use the script \href{https://github.com/huggingface/diffusers/blob/main/examples/dreambooth/train_dreambooth_lora.py}{\texttt{train\_dreambooth\_lora.py}} \cite{dreambooth} with the diffusers library \cite{github-diffusers}.
we use the standard recipe to fine-tune the models in all ranges\cite{dreambooth_readme} see tab\ref{table:fine_tune_hyperparams}. we use batch size 8 for range 1-1000 for computational resources and 1000 training step. in the ablations we don't change any hyper-parameter except the ablate one.

Each model took approximately 30-50 minutes to fine-tune. We used GPUs with 16-21GB of RAM, such as the NVIDIA RTX A5000. The DSiRe process, however, does not require GPUs and can run on CPUs.

\begin{table*}[h!]
    \begin{minipage}{.40\linewidth}
        \caption{\textit{\textbf{ranges 1-6 and 1-50}}}
        \centering
        \begin{tabular}{l@{\hskip5pt}l@{\hskip5pt}}    
             Name & Value \\
             \midrule
             \texttt{lora\_rank} ($r$) & $r$ \\
             \texttt{lr} & $1e-4$ \\
             \texttt{batch\_size} & $1$ \\
             \texttt{gradient\_accumulation\_steps} & $1$ \\
             \texttt{learning\_rate\_scheduler} & Constant \\
             \texttt{training\_steps} & $1400$ \\
             \texttt{warmup\_ratio} & $0$ \\
             \texttt{dataset} & \begin{tabular}[l]{@{}l@{}}\texttt{imagenet}\cite{imagenet} \\ concept101\cite{dataset_multi} \end{tabular}\\
             \texttt{seeds} & $0$ \\
        \end{tabular}
        \label{table:mistral_sft_hyperparams}
    \end{minipage}%
    \hfill
    \begin{minipage}{.4\linewidth}
        \caption{\textit{\textbf{range 1-1000 Hyper-parameters}}}
        \centering
        \begin{tabular}{l@{\hskip5pt}l@{\hskip5pt}}    
             Name & Value \\
             \midrule
             \texttt{lora\_rank} ($r$) & $32$ \\
             \texttt{lr} & $1e-4$ \\
             \texttt{batch\_size} & $8$ \\
             \texttt{gradient\_accumulation\_steps} & $1$ \\
             \texttt{learning\_rate\_scheduler} & Constant \\
             \texttt{training\_steps} & $1000$ \\
             \texttt{warmup\_ratio} & $0$ \\
             \texttt{dataset} & \texttt{imagenet} \\
             \texttt{seeds} & $0$ \\
        \end{tabular}
        \label{table:fine_tune_hyperparams}
    \end{minipage} 
\end{table*}

\paragraph{Experiment Settings.}
In addition to the experiment settings described in Section~\ref{sec:experiments}, we used the following configurations for our models:

- For models in the ranges 1-6 and 1-50, we used the checkpoint at iteration 1200.
- For models in the range 1-1000, we used the checkpoint at iteration 1000.

We used a fixed seed of 42 to split the train and test data for every experiment.

\paragraph{Layer weigh matrices}
In line with our analysis see Sec.\ref{sec:pre_analysis}, given $n$ example weights for each $A_i,B_i$ we wish to build a separate classifier for each one. Knowing the relation between the singular values and the dataset size, we decompose each matrix using the singular value decomposition (SVD), and use the ordered set of singular values as features for our classifiers. Formally, we note the singular values of $A_{ij}$ as $\Sigma_{A_{ij}}$ and the singular values of $B_{ij}$ as $\Sigma_{B_{ij}}$. 
We include the singular values of $B_{ij} \cdot A_{ij}$ denoted as $\Sigma_{B_{ij} \cdot A_{ij}}$. 
Additionally, our observations indicate that the product $B_i \cdot A_i$ also provides useful information for data size recovery. 
Thus, for each LoRA matrix, we obtain a dataset with $n$ samples, where each sample is a vector of singular values $\Sigma_{ij}$, paired with a corresponding label $y_j$. Our method then trains three separate kNN-classifiers with $K=1$ for each layer over (i) $A_i$ (ii) $B_i$ and (iii) $B_iA_i$. At inference time, the predictions from all classifiers are merged by majority voting.